\documentclass{article}

\usepackage{PRIMEarxiv}

\usepackage[utf8]{inputenc} 
\usepackage[T1]{fontenc}    
\usepackage{hyperref}       
\usepackage{url}            
\usepackage{booktabs}       
\usepackage{amsfonts}       
\usepackage{nicefrac}       
\usepackage{microtype}      
\usepackage{fancyhdr}       
\usepackage{graphicx}       
\graphicspath{{media/}}     

\usepackage{graphicx}
\usepackage{multirow} 
\usepackage{multicol} 
\usepackage{colortbl}
\usepackage{array}
\usepackage{makecell}
\usepackage{amsmath}
\usepackage{amssymb}
\usepackage[export]{adjustbox}
\usepackage[table]{xcolor}
\usepackage{subcaption}
\usepackage[numbers]{natbib}
\usepackage{hyperref}

\definecolor{lightgreen}{RGB}{230,255,230}
\definecolor{lightred}{RGB}{255,230,230}

\usepackage{booktabs}
\usepackage{tabularx}
\newcolumntype{Y}{>{\centering\arraybackslash}X}
\usepackage{multirow}

\usepackage{stackengine}

\title{Neuro-Symbolic Scene Graph Conditioning for Synthetic Image Dataset Generation
}

\author{
  Giacomo Savazzi, Eugenio Lomurno, Cristian Sbrolli, Agnese Chiatti, Matteo Matteucci \\
  Politecnico di Milano \\
  Department of Electronics, Information and Bioengineering\\
  Via Ponzio 34/5, 20133 Milan, Italy\\
  \texttt{\{giacomo.savazzi\}@mail.polimi.it} \\
  \texttt{\{eugenio.lomurno, cristian.sbrolli\}@polimi.it} \\
  \texttt{\{agnese.chiatti, matteo.matteucci\}@polimi.it} \\
}

\begin{document}
\maketitle

\begin{abstract}
\noindent As machine learning models increase in scale and complexity, obtaining sufficient training data has become a critical bottleneck due to acquisition costs, privacy constraints, and data scarcity in specialised domains. While synthetic data generation has emerged as a promising alternative, a notable performance gap remains compared to models trained on real data, particularly as task complexity grows. Concurrently, Neuro-Symbolic methods, which combine neural networks' learning strengths with symbolic reasoning's structured representations, have demonstrated significant potential across various cognitive tasks.
This paper explores the utility of Neuro-Symbolic conditioning for synthetic image dataset generation, focusing specifically on improving the performance of Scene Graph Generation models. The research investigates whether structured symbolic representations in the form of scene graphs can enhance synthetic data quality through explicit encoding of relational constraints. 
The results demonstrate that Neuro-Symbolic conditioning yields significant improvements of up to +2.59\% in standard Recall metrics and +2.83\% in No Graph Constraint Recall metrics when used for dataset augmentation. These findings establish that merging Neuro-Symbolic and generative approaches produces synthetic data with complementary structural information that enhances model performance when combined with real data, providing a novel approach to overcome data scarcity limitations even for complex visual reasoning tasks.
\end{abstract}

\keywords{Synthetic Data Generation  \and Neuro-Symbolic AI \and Scene Graph Generation \and Data Augmentation \and Visual Reasoning.}

\section{Introduction}
\noindent Recently, research on synthetic dataset generation has emerged as a critical direction for advancing deep learning architectures. As models continue to scale in complexity and parameter count, acquiring and annotating large-scale real-world datasets remains prohibitively resource-intensive due to costs, privacy regulations, and domain-specific data scarcity.
Synthetic datasets offer a scalable alternative with reduced annotation overhead and controllable attribute distribution. Despite these advantages, models trained exclusively on synthetic data consistently underperform compared to those trained on real datasets, particularly on high-level reasoning tasks. This performance gap widens further as task complexity increases, even when scaling dataset size.
To address these limitations, various conditioning techniques for generative models have been explored.

\noindent Simultaneously, Neuro-Symbolic (NeSy) approaches have gained traction in the machine learning community~\cite{sarker2022neuro}. These methods integrate neural representation learning with symbolic reasoning frameworks, combining data-driven optimization with structured knowledge. While NeSy architectures have demonstrated enhanced generalization in reasoning tasks, their potential for improving synthetic data generation remains largely unexplored.
Scene graphs—structured representations where nodes represent objects and edges represent relations—offer a promising symbolic framework for conditioning image generation. 

\noindent This work explores the integration of such NeSy approaches into dataset generation to improve performance on complex tasks, particularly Scene Graph Generation (SGG) from images. The hypothesis posits that scene graphs encode useful background knowledge that can guide the generation process, ensuring synthetic data adhere to structural and semantic constraints. To the best of current knowledge, this represents the first systematic exploration of NeSy conditioning for synthetic dataset generation in the context of SGG tasks.
The experimental framework employs SGAdapter~\cite{SGAdapter} for Scene Graph-to-Image generation, with Stable Diffusion 2.0~\cite{ho2020denoising} as the baseline model. CausalTDE~\cite{CausalTDE} serves as the SGG evaluation model. 

\noindent The key contributions of this work are threefold: (1) proposing a novel framework that integrates scene graph-based symbolic knowledge into synthetic dataset generation; (2) demonstrating that structurally-guided generation produces complementary training signals that enhance augmentation efficacy despite lower perceptual fidelity; and (3) providing empirical evidence that NeSy conditioning addresses specific limitations in conventional synthetic data generation for complex visual reasoning tasks.

\section{Related Works}
\noindent The creation of effective synthetic datasets for augmentation and model training necessitates generative architectures capable of high-quality image synthesis. Among these, Generative Adversarial Networks (GAN) and Denoising Diffusion Probabilistic Models (DDPM) have emerged as predominant families of methods~\cite{goodfellow2014generative, ho2020denoising}. Despite their distinct operational principles, both approaches can be conditioned in various ways to produce images that correspond to specific domains or visual concepts.

\subsection{Conditioning Generative Models}\label{subsec:conditioning_methods}
\noindent The evolution of GANs quickly led to conditional formulations, allowing the generation process to be guided by supplementary information provided to both generator and discriminator components~\cite{mirza2014conditional}. Particularly promising has been the conditioning on class labels, which offers benefits through training set expansion and noise input variance truncation, ultimately enhancing the quality of generated images~\cite{brock2018large}.

\noindent The landscape of generative models has been further enriched by the exploration of text-prompt conditioning. Ku \textit{et al.} advanced this field by developing a regressor that produces refined text-conditioned vectors, enabling nuanced control over subtle features in generated images~\cite{ku2023textcontrolgan}. Complementary approaches harnessed CLIP's comprehensive understanding of visual scenes, simultaneously reducing training requirements while elevating synthesis quality~\cite{tao2023galip, radford2021learning}.

\noindent In the realm of diffusion models, textual prompts are similarly employed, often alongside guidance strategies that carefully balance fidelity with diversity. Nichol \textit{et al.} demonstrated the superiority of classifier-free guidance—achieved by blending predictions with and without text conditioning—over approaches relying on CLIP guidance~\cite{nichol2021glide}. The widely-used Stable Diffusion implements this concept through its Guidance Scale parameter, which regulates how closely generated images adhere to provided text prompts~\cite{rombach2022high}. 

\subsection{Learning from Synthetic Data}
\label{subsec:learning_from_synthetic_data}
\noindent The research trajectory has recently shifted from generating individual synthetic samples toward producing comprehensive synthetic datasets, prioritising collective diversity over the visual quality of individual images.

\noindent This transition has sparked numerous studies focused on enhancing the diversity of synthetic datasets. Researchers have discovered that minimal prompt engineering combined with reduced guidance scale in Stable Diffusion can yield performance comparable to that achieved with real data~\cite{sariyildiz2023fake}. The critical importance of the guidance scale parameter for synthetic image generation has been found to benefit downstream classification tasks~\cite{lomurno2024stable}. A parallel finding revealed that expanding input noise variance enhances the informativeness of GAN-generated datasets, even when this approach might compromise the visual fidelity of individual samples~\cite{lampis2023bridging}.

\noindent Shipard \textit{et al.} introduced an innovative approach based on multiple textual variations, amalgamated into a "Bag of Tricks" that effectively broadens synthetic dataset diversity. Taking a different approach, Lei \textit{et al.} employed captioning models to craft detailed prompts, combining image descriptions with class labels to enhance the distinction between foreground and background elements~\cite{lei2023image}.

\noindent Complementing these generation-focused methods, researchers have explored post-generation enrichment techniques. The Generative Knowledge Distillation technique elegantly leverages models trained on real data (Teacher) to produce nuanced soft labels for synthetic datasets, which then serve to train new models (Student)~\cite{lomurno2025synthetic}. This approach has demonstrated remarkable effectiveness in enhancing classification performance while increasing resilience against inference attacks in both local and federated learning environments~\cite{lomurno2025federated}.

\subsection{Neuro-Symbolic Artificial Intelligence}\label{subsec:neuro_symbolic_ai}
\noindent Neuro-Symbolic (NeSy) AI represents an innovative approach that integrates neural and symbolic paradigms in Artificial Intelligence. This integration strategically combines the perceptual strengths of neural systems with the reasoning capabilities of symbolic approaches, creating a synergy that aims at addressing the limitations of both paradigms.

\noindent Neural systems excel at processing unstructured data and learning patterns through approximation via deep neural networks. Their strength lies in perception tasks such as image recognition and automatic information extraction from vast datasets. However, they typically function as black boxes, unable to provide transparent explanations for their operations and outputs.

\noindent In contrast, symbolic systems operate on structured data like logic rules or knowledge graphs, employing search algorithms and logical reasoning. These systems demonstrate superior performance in cognitive tasks and offer greater explainability but require specialized domain knowledge to function effectively. This domain-specific knowledge is typically expensive to acquire and it can also include procedural, tacit, and commonsense constraints that are hard to formalise explicitly as we take them for granted when describing images (e.g., inanimate objects do not float mid-air unless supported by other surfaces). 

\noindent The survey by Yu et al.~\cite{yu2023survey} categorizes NeSy systems into three principal integration approaches: Learning for Reasoning, where neural systems enhance symbolic reasoning by reducing search spaces or extracting meaningful symbols; Reasoning for Learning, where symbolic knowledge guides neural learning to improve performance and interpretability; and Learning-Reasoning, which establishes bidirectional interaction between both paradigms to maximize their combined problem-solving capabilities.

\subsection{Scene Graphs and Scene Graph-to-Image Generation}\label{subsec:scene_graphs}
\noindent Scene graphs provide a structured representation of visual scenes, where nodes represent objects and edges define the relationships between object instances. Since 2018, these structured representations have emerged as a more structured representation than natural language descriptions to use as input modality for image generation.

\noindent Natural language suffers from ambiguity when describing multiple objects and their spatial or functional relationships, being heavily dependent on syntax and semantic interpretation. Furthermore, research has revealed that CLIP~\cite{radford2021learning} text encoders, commonly used in conditioning generative models, tend to underweight the relationships between objects and their attributes. Hence, since scene graphs offer precise structural representations that explicitly define object relationships, we argue that they could provide valuable complementary information to enhance the conditioning of generative models.

\noindent The field of scene graph-to-image generation gained momentum following the pioneering work of Johnson et al.~\cite{SG2Im}, who introduced a two-step approach: first converting the scene graph into an image layout, then using this layout to condition a GAN-based generative model. Subsequent research~\cite{SGlayout,SceneGenie,SGTransformer} refined this methodology, using scene graphs to generate layouts, semantic maps, or bounding boxes that subsequently guide the generation process.

\noindent While this two-step approach provided robust structural conditioning, it significantly limited diversity and increased computational demands during inference. Consequently, recent research~\cite{SGAdapter,SGDiff} has shifted toward directly conditioning the diffusion process of Latent Diffusion Models (LDMs) with scene graph information, offering a more balanced approach to structured image generation.

\subsubsection{Stable Diffusion}\label{subsubsec:stable_diffusion}
\noindent Stable Diffusion 2, developed by Rombach et al.~\cite{ho2020denoising}, represents a state-of-the-art LDM for image generation. It implements the reverse diffusion process through a U-Net network trained to predict the noise $\epsilon$ added at a randomly sampled step $t$. The model can be conditioned on an input $y$ using a denoising autoencoder $\epsilon_\theta \left(z_t,t,y \right)$ and mapping the encoded conditioning input $\tau_\theta\left(y\right)\ \in\mathbb{R}^{M\ \times\ d_r}$ to the U-net's intermediate layers via cross-attention. The conditional LDM is learned through the loss function
\begin{equation}
    \scalebox{0.85}{$
      L_{LDM}=\mathbb{E}_{\varepsilon(x),y,\epsilon\sim N(0,1),t}\Bigl[\left\|\epsilon-\epsilon_\theta\left(z_t,t,\tau_\theta(y)\right)\right\|^2_2\Bigr].
  $}
\end{equation}

\subsubsection{SGAdapter}\label{subsubsec:sgadapter}
\noindent Shen et al.~\cite{SGAdapter} proposed SGAdapter, an innovative model that enhances Stable Diffusion's conditioning mechanism by incorporating scene graph information into the text embeddings. This adapter addresses a fundamental limitation in the CLIP text encoder's attention mechanism, which compromises semantic structure when processing multiple relations by masking subsequent tokens and computing attention only with preceding ones.

\noindent Rather than undertaking the resource-intensive task of retraining the entire CLIP model, SGAdapter enriches the CLIP text embedding $w = E_T(c), w \in \mathbb{R}^{N\times D}$ with scene graph information through a specialized attention mechanism. This process uses a unified graph embedding $e \in \mathbb{R}^{K\times D}$ derived from the concatenated embeddings of each relation in the graph, computing attention as:
\begin{equation}
    \scalebox{0.85}{$
    w' = Attention\left(Q_w,K_e,V_e,M^{sg}\right),
    $}
\end{equation}
where the scene graph attention mask is defined as:
\begin{equation}
    \scalebox{0.85}{$
    M^{sg}_{ik} = \begin{cases}0 &if\ \tau(i)=k, \\ -\infty &otherwise. \end{cases}
    $}
\end{equation}

\noindent The function $\tau(i)$ maps token $i$ to its corresponding relation.

\subsection{Scene Graph Generation}\label{subsec:scene_graph_generation}
\noindent Scene Graph Generation (SGG) represents one of the most challenging tasks in image analysis, with significant implications for downstream applications such as Visual Question Answering. The SGG process typically follows a two-step approach: first detecting and recognizing the principal objects in an image, then identifying the relationships between these objects.

\noindent Several sophisticated models have emerged as state-of-the-art in SGG, including Neural-Motifs~\cite{Motifs}, VCTree~\cite{VCTree}, SpeaQ~\cite{SpeaQ}, and CausalTDE~\cite{CausalTDE}, each employing distinct strategies to improve accuracy in the relation prediction task.

\subsubsection{CausalTDE}\label{subsubsec:causaltde}
\noindent Tang et al.~\cite{CausalTDE} introduced CausalTDE, a model-agnostic approach that integrates Causal Total Direct Effect analysis into SGG frameworks. Widely recognized as a state-of-the-art framework for scene graph generation, CausalTDE has established itself as a standard in the field due to its superior performance and computational efficiency. This innovative methodology addresses the negative biases in existing SGG models arising from predicate frequency imbalances in training data, while maintaining beneficial biases that help filter out unrealistic predictions.

\noindent The CausalTDE approach first trains a biased SGG framework, then applies Total Direct Effect analysis to its predictions. This involves generating both a biased prediction $Y_x(U)$ and a counterfactual prediction $Y_{\bar{x},z}(u)$ that primarily depends on context-specific biases. The TDE is calculated as the difference between these predictions. Formally:
\begin{equation}
    \scalebox{0.85}{$
    TDE = Y_x(U) - Y_{\bar{x},z}(u),
    $}
\end{equation}
resulting in an unbiased relation prediction that significantly improves SGG performance.

\section{Methods}
\noindent This research investigates the efficacy of Neuro-Symbolic conditioning in synthetic image generation for dataset augmentation. The work presents a systematic analysis of how structured semantic information in the form of subject-predicate-object triples (e.g., a bike near a house) can enhance model performance when integrated with real-world data. We achieve this integration by conditioning synthetic image generation models with scene graphs. 
The investigation specifically targets Scene Graph Generation (SGG) tasks, providing a comparative analysis between conventional conditioning methodologies and Neuro-Symbolic approaches, representing, to the best of our knowledge, the first systematic study of this integration in the literature.

\subsection{Neuro-Symbolic Conditioning Framework}
\noindent The proposed methodology employs SGAdapter~\cite{SGAdapter} as a conditioning mechanism integrated with Stable Diffusion 2.0~\cite{ho2020denoising}. SGAdapter functions as an adaptation layer that incorporates structured scene graph information into the diffusion process through specialised attention mechanisms, rather than operating as an independent generative model.

\noindent This conditioning architecture implements two distinct attention mechanisms:
\begin{itemize}
    \item \textbf{Scene Graph Cross-Attention (SGC-Att)}: This mechanism directly integrates scene graph information, represented as triples of subject-relation-object, into the textual embedding by computing cross-attention between the triples embedding and SGAdapter's CLIP-based text embedding. Through a specialised attention mask, textual tokens are matched with corresponding relations in the scene graph.
    \item \textbf{Self-Attention (S-Att)}: This module implements a relational attention masking mechanism that restricts token interactions to only those sharing direct relational connections. For example, in "A man beside his car and in front of the house with a tree next to it," "A man" attends only to its direct relational elements ("beside his car" and "in front of the house") while masking unrelated tokens.
    
\end{itemize}

\noindent Four distinct Neuro-Symbolic conditioning configurations are examined:
\begin{itemize}
    \item \textbf{Configuration 1}: Utilising both SGC-Att and S-Att masks simultaneously.
    \item \textbf{Configuration 2}: Implementing only the SGC-Att mask.
    \item \textbf{Configuration 3}: Employing exclusively the S-Att mask.
    \item \textbf{Configuration 4}: Operating without either attention mask, using BLIP~2~\cite{li2023blip} for generating semantically rich textual captions. This approach is adopted as the absence of attention masks eliminates the need for maintaining tensor format consistency for attention operations. 
\end{itemize}
\noindent The SGC-Att mechanism is computed in all configurations, regardless of whether the SGC-Att mask is applied, whereas the S-Att computation is entirely skipped in configurations where the S-Att mask is not provided as input.

\noindent It is also considered the standard Stable Diffusion 2.0 without any Neuro-Symbolic conditioning as the baseline, to directly compare the performance of conventional text-conditioned generation with the proposed Neuro-Symbolic approach.

\subsection{Synthetic Data Generation and Annotation Protocol}
The synthetic dataset generation methodology comprises two critical phases: controlled image synthesis and subsequent annotation extraction.

\noindent For image synthesis, both the baseline Stable Diffusion 2.0 and its four configurations with SGAdapter produce synthetically-generated images. The four SGAdapter-enhanced configurations generate images from captions and scene graph descriptions, while standard Stable Diffusion 2.0 generates images directly from captions alone. Regardless of the generator employed, images are produced in batches of 8 with a resolution of $512\times512$ and 50 diffusion steps.

\noindent The annotation extraction process employs a hybrid approach that integrates generated annotations with object detection predictions. Specifically, for each relation defined during the generation process, the system verifies the existence of corresponding object pairs within the detection output. When valid object pairs are identified, their semantic relationships are incorporated into the annotation set. This annotation extraction is performed using the pre-trained CausalTDE model, which is based on Faster RCNN for object prediction, with a confidence threshold of $0.3$ applied both to boundig box in relation to the class predictions, to filter generated annotations.

\subsection{Evaluation Methodology}
\noindent The evaluation framework primarily focuses on dataset augmentation efficacy, with complementary analysis of models trained exclusively on synthetic data. Two main experimental scenarios are examined:
\begin{itemize}
    \item \textbf{Dataset Augmentation}: Integration of synthetically generated images with real-world data to enhance model performance and robustness.
    \item \textbf{Training from Synthetic Data}: Models trained exclusively on synthetic images.
\end{itemize}

\noindent CausalTDE~\cite{CausalTDE} applied to MotifsNet~\cite{Motifs} serves as reference architecture for the evaluation, due to its state-of-the-art performance characteristics and computational efficiency in the Scene Graph Generation task. Performance assessment utilises a comprehensive suite of metrics:
\begin{itemize}
    \item \textbf{Recall@K with K$\in[20,50,100]$}: Quantifies the proportion of correct predictions within the top K candidates.
    \item \textbf{No Graph Constraint Recall@K with K$\in[20,50,100]$}: Extends the Recall@K evaluation by considering semantically plausible relationships beyond ground truth annotations.
\end{itemize}

\noindent This evaluation protocol enables a detailed analysis of how Neuro-Symbolic conditioning affects both specific relation prediction accuracy as well as the overall model generalisation capabilities in the context of data augmentation.

\section{Experimental Setup}
\noindent The dataset used for the experiments in this paper is Visual Genome~\cite{VisualGenome}, which contains dense annotations of image descriptions, objects, attributes, and relationships. Visual Genome comprises 108,000 images, including 75,000 objects and 35,000 predicates.

\noindent Training SGAdapter requires as input both image captions and the mapping of relations to caption tokens. Several data preparation steps were necessary for the Visual Genome dataset, as not all images therein were suitable for the intended purpose. The following filtering criteria were therefore applied:
\begin{itemize}
\item Images with a width or height smaller than 500 pixels are excluded, as they would require excessive upscaling factors, potentially introducing significant visual artifacts.
\item Images with bounding boxes smaller than 32 pixels in width or height are excluded, as such minuscule objects could be inadequately represented during generator training, potentially leading to recognition failures and inconsistent generation results.
\item Objects and attributes that appear less than 3 and 10 times respectively are excluded, as their rare occurrence hinders their utility for training purposes.
\item Images without any annotated relationships or objects are excluded, since, in our experimental setup, having access to a scene graph is a pre-requisite for training SGAdapter.
\item Images with more than 20 relationships are excluded, since SGAdapter cannot effectively process predicate sets exceeding this threshold.
\end{itemize}

\noindent Captions are then created by concatenating objects with their attributes, followed by relations to facilitate the construction of relation-token mappings, thus avoiding the use of captioners or human effort. Captions exceeding 77 tokens are excluded from the dataset since SGAdapter truncates longer sequences, potentially disrupting relationship integrity. Ultimately, the relation-tokens mappings are systematically generated by associating each relation with its corresponding token indices in the caption.

\noindent The raw Visual Genome dataset proves unsuitable for training CausalTDE, as 92\% of the 35,000 predicates have no more than ten instances. Therefore, the widely used Visual Genome split is used, which contains only the most frequent 150 objects and 50 predicates, in line with Tang et al.~\cite{CausalTDE}.

\noindent Visual Genome split data annotations are utilized to generate synthetic datasets, with captions and mappings generated through the same process and filtering step described above.

\noindent A distinct synthetic dataset is generated for each configuration of SGAdapter and for the standalone Stable Diffusion 2.0 model. These synthetic datasets are subsequently used to augment the real dataset in training a CausalTDE model from scratch. For evaluation purposes, a test set comprising 26,000 images extracted from Visual Genome is employed. Additionally, experiments are conducted to measure SGG performance when purely synthetic data is used in place of the real dataset.

\noindent Configuration 4, as described in the Methods section, utilizes neither of the masks in both attention mechanisms and, since it does not require token-to-relation mapping, employs realistic textual captions extracted using BLIP-2~\cite{li2023blip}. As Visual Genome does not provide image captions for all images (only those images originally from the COCO dataset, which Visual Genome augments with subject-relationship-object annotations in the form of scene graphs), BLIP-2 was employed for caption generation.

\noindent The BLIP-2 model, which uses the OPT large language model with 2.7 billion parameters, is selected for image captioning. This specific checkpoint\footnote{\scriptsize\url{https://huggingface.co/Salesforce/blip2-opt-2.7b}} is chosen due to its comparatively lower computational requirements. The pre-trained BLIP-2 model is implemented directly without any additional fine-tuning on the target dataset, with the sole hyper-parameter modification being \texttt{max\_new\_tokens=20} to effectively control caption length.

\noindent All four configurations of the generative model are trained for 400 steps with a consistent batch size of 8. The CausalTDE model is trained using the parameters precisely outlined by Tang et al.~\cite{CausalTDE}. Since CausalTDE systematically filters images without annotated relations, the size of synthetic datasets is standardized to 20,000 images across all configurations.

\noindent All experiments presented so far are conducted on a pair of NVIDIA Quadro RTX 6000 graphics processing units.

\section{Results}
\noindent The experiments yield interesting results in both scenarios: when augmenting the dataset and when training exclusively on synthetic data, providing insights on the effects of neuro-symbolic approaches in image generation. When used for augmentation, the neuro-symbolically generated dataset clearly outperforms the baseline-generated ones. However, when training solely on synthetic data, an unexpected pattern emerges: the Stable Diffusion 2.0 generated dataset surprisingly outperforms the neuro-symbolic ones.

\subsection{Dataset Augmentation}\label{DatasetAug}
\noindent In the Dataset Augmentation experiment, the synthetic datasets effectively augment the real dataset, as highlighted by the improved Recall@K performance (Table~\ref{tab:AugRecall}). Moreover, the neuro-symbolically generated images produce a substantially higher performance increase compared to those from Stable Diffusion 2.0. Specifically, SGAdapter Configuration 1 achieves an average Recall@K improvement of $2.59\%$ versus the $0.36\%$ obtained with Stable Diffusion 2.0.

\begin{table*}[t]
    \centering
    \scriptsize
    \caption{\footnotesize Performance comparison of Recall and No Graph Constraint Recall metrics evaluated on the real data test set. The "Real Dataset" row shows results for the CausalTDE model trained exclusively on real data, while other rows represent the same model trained on various augmented datasets. Green cells indicate performance equal to or better than the baseline (real data only training), while red cells show performance degradation. The Mean $\Delta$ columns display the average performance change relative to the Real Dataset baseline.}
    \resizebox{\linewidth}{!}{
        \begin{tabular}{|c|c|c|c|c|c|c|c|c|c|c|}
            \hline
            \multirow{2}{*}{\textbf{Model}} & \multicolumn{2}{c|}{\textbf{Attention Mask}} & \multicolumn{4}{c|}{\textbf{Recall}} & \multicolumn{4}{c|}{\makecell{\textbf{No Graph Constraint} \\ \textbf{Recall}}} \\
            \cline{2-3}\cline{4-11}
            & SGC-Att & S-Att & R@20 & R@50 & R@100 & \makecell{Mean $\Delta$} & NG-R@20 & NG-R@50 & NG-R@100 & \makecell{Mean $\Delta$} \\
            \hline
            \hline
            \makecell{Real \\ Dataset} & / & /& 0.0983 & 0.1428 & 0.1763 & - & 0.1039 & 0.1619 & 0.2125 & - \\
            \hline
            \hline
            \makecell{Stable \\ Diffusion 2.0} & /& /  & \cellcolor{lightgreen}0.1022 & \cellcolor{lightgreen}0.1469 & \cellcolor{lightgreen}0.1791 & \cellcolor{lightgreen}0.0036 & \cellcolor{lightgreen}0.1108 & \cellcolor{lightgreen}0.1706 & \cellcolor{lightgreen}0.2223 & \cellcolor{lightgreen}0.0085 \\
            \hline
            \hline
            \makecell{SGAdapter\\Configuration 1} & \checkmark & \checkmark & \cellcolor{lightgreen}\textbf{0.1200} & \cellcolor{lightgreen}\textbf{0.1695} & \cellcolor{lightgreen}\textbf{0.2057} & \cellcolor{lightgreen}\textbf{0.0259} & \cellcolor{lightgreen}\textbf{0.1278} & \cellcolor{lightgreen}\textbf{0.1912} & \cellcolor{lightgreen}\textbf{0.2441} & \cellcolor{lightgreen}\textbf{0.0283} \\
            \hline 
            \makecell{SGAdapter\\Configuration 2} & \checkmark & $\times$ & \cellcolor{lightgreen}0.1102 & \cellcolor{lightgreen}0.1599 & \cellcolor{lightgreen}\underline{0.1968} & \cellcolor{lightgreen}0.0165 & \cellcolor{lightgreen}0.1185 & \cellcolor{lightgreen}\underline{0.1834} & \cellcolor{lightgreen}\underline{0.2379} & \cellcolor{lightgreen}0.0205 \\
            \hline
            \makecell{SGAdapter\\Configuration 3} & $\times$ & \checkmark  & \cellcolor{lightgreen}0.1065 & \cellcolor{lightgreen}0.1545 & \cellcolor{lightgreen}0.1907 & \cellcolor{lightgreen}0.0114 & \cellcolor{lightgreen}0.1179 & \cellcolor{lightgreen}0.1812 & \cellcolor{lightgreen}0.2351 & \cellcolor{lightgreen}0.0186 \\
            \hline
            \makecell{SGAdapter\\Configuration 4} & $\times$ & $\times$ & \cellcolor{lightgreen}\underline{0.1150} & \cellcolor{lightgreen}\underline{0.1609} & \cellcolor{lightgreen}0.1946 & \cellcolor{lightgreen}\underline{0.0177} & \cellcolor{lightgreen}\underline{0.1215} & \cellcolor{lightgreen}0.1826 & \cellcolor{lightgreen}0.2345 & \cellcolor{lightgreen}\underline{0.0201} \\
            \hline
        \end{tabular}
    }
    \label{tab:AugRecall}
\end{table*}

\noindent These results indicate that neuro-symbolically generated images provide more valuable information when combined with the real dataset than those generated by Stable Diffusion 2.0. The greater performance improvement suggests NeSy-generated images contribute semantic content that enhances the dataset's overall informativeness, enabling more effective learning.

\noindent Further analysis of the Recall@100 for each predicate, Table~\ref{tab:AugPredRecall}, shows a particular behaviour in the augmented datasets, which tend to increase or perform similarly to the original one for most predicates, but for a few of them they cause performance deterioration. This behaviour also changes between the augmented datasets, as some of them perform well while others cause performance drop.

\noindent These results are probably caused by the inability of the generators to represent some predicates, thus, the badly generated images impair the learning ability of the SGG model. The generators learn to represent different predicates, leading to different performances among them. This may cause the model to specialize towards the predicates that the generator can properly represent. 

\begin{table}[t]
    \centering
    \caption{\footnotesize Recall@100 for each predicate of the trained model, comparing performance when trained on augmented datasets versus the Real Data baseline, all evaluated on the same test set. Green cells indicate values equal to or exceeding the Real Data baseline, whereas red cells indicate lower values. Predicates with 0 Recall across all models have been omitted.}
    \resizebox{\textwidth}{!}{
        \begin{tabular}{|c||c||c||c|c|c|c|}
            \hline
             \textbf{Predicate} & \textbf{Real Data} & \textbf{Stable Diffusion 2.0} & \makecell{\textbf{SGAdapter} \\ \textbf{Configuration 1}}  & \makecell{\textbf{SGAdapter} \\ \textbf{Configuration 2}} & \makecell{\textbf{SGAdapter} \\ \textbf{Configuration 3}} & \makecell{\textbf{SGAdapter} \\ \textbf{Configuration 4}}  \\   
             \hline
           above & \underline{0.0890} &    \cellcolor{lightgreen}\textbf{0.0933} & \cellcolor{lightred}0.0406 & \cellcolor{lightred}0.0556 & \cellcolor{lightred}0.0683 & \cellcolor{lightred}0.0494 \\ \hline
            along & \textbf{0.0887} & \cellcolor{lightred}0.0550 & \cellcolor{lightred}0.0474 & \cellcolor{lightred}\underline{0.0726} & \cellcolor{lightred}0.0596 & \cellcolor{lightred}0 \\ \hline
            and & \underline{0.0118} & \cellcolor{lightgreen}\textbf{0.0237} & \cellcolor{lightred}0 & \cellcolor{lightred}0 & \cellcolor{lightred}0 & \cellcolor{lightred}0 \\ \hline
            at & 0.2233 & \cellcolor{lightred}0.2173 & \cellcolor{lightgreen}0.2442 & \cellcolor{lightgreen}\textbf{0.2599} & \cellcolor{lightgreen}\underline{0.2459} & \cellcolor{lightgreen}0.2431 \\ \hline
            attached to & \textbf{0.0539} & \cellcolor{lightred}\underline{0.0325} & \cellcolor{lightred}0.0078 & \cellcolor{lightred}0.0097 & \cellcolor{lightred}0.0008 & \cellcolor{lightred}0.0063 \\ \hline
            behind & 0.2323 & \cellcolor{lightred}0.2246 & \cellcolor{lightgreen}0.2443 & \cellcolor{lightgreen}\textbf{0.2547} & \cellcolor{lightgreen}\underline{0.2491} & \cellcolor{lightgreen}0.2480 \\ \hline
            belonging to & \underline{0.1093} & \cellcolor{lightgreen}\textbf{0.1780} & \cellcolor{lightred}0 & \cellcolor{lightred}0.0048 & \cellcolor{lightred}0 & \cellcolor{lightred}0 \\ \hline
            carrying & 0.2068 & \cellcolor{lightgreen}0.2165 & \cellcolor{lightgreen}\textbf{0.2495} \cellcolor{lightgreen}& \cellcolor{lightgreen}0.2278 & \cellcolor{lightgreen}0.2183 & \cellcolor{lightgreen}\underline{0.2280} \\ \hline
            eating & \underline{0.0889} & \cellcolor{lightgreen}\textbf{0.0920} & \cellcolor{lightred}0.0199 & \cellcolor{lightred}0 & \cellcolor{lightred}0 & \cellcolor{lightred}0 \\ \hline
            for & 0.0123 & \cellcolor{lightred}0.0012 & \cellcolor{lightred}0 & \cellcolor{lightred}0 & \cellcolor{lightgreen}\textbf{0.0389} & \cellcolor{lightgreen}\underline{0.0369} \\ \hline
            hanging from & \textbf{0.0378} & \cellcolor{lightred}\underline{0.0120} & \cellcolor{lightred}0 & \cellcolor{lightred}0 & \cellcolor{lightred}0 & \cellcolor{lightred}0 \\ \hline
            has & 0.3421 & \cellcolor{lightgreen}0.3493 & \cellcolor{lightgreen}\underline{0.3700} & \cellcolor{lightgreen}\textbf{0.3774} & \cellcolor{lightgreen}0.3675 & \cellcolor{lightgreen}0.3695 \\ \hline
            holding & 0.1738 & \cellcolor{lightgreen}0.1859 & \cellcolor{lightgreen}\textbf{0.2038} & \cellcolor{lightred}0.1583 & \cellcolor{lightgreen}\underline{0.1896} & \cellcolor{lightred}0.1665 \\ \hline
            in & \textbf{0.1082} & \cellcolor{lightred}0.0977 & \cellcolor{lightred}0.1013 & \cellcolor{lightred}0.0973 & \cellcolor{lightred}\underline{0.1050} & \cellcolor{lightred}0.0988 \\ \hline
            in front of & 0.1737 & \cellcolor{lightred}0.1524 & \cellcolor{lightgreen}\underline{0.1755} & \cellcolor{lightred}0.1653 & \cellcolor{lightgreen}\textbf{0.1914} & \cellcolor{lightred}0.1351 \\ \hline
            laying on & \textbf{0.0913} & \cellcolor{lightred}\underline{0.0045} & \cellcolor{lightred}0 & \cellcolor{lightred}0 & \cellcolor{lightred}0 & \cellcolor{lightred}0 \\ \hline
            looking at & \textbf{0.1329} & \cellcolor{lightred}0.1058 & \cellcolor{lightred}0.0407 & \cellcolor{lightred}0.0789 & \cellcolor{lightred}\underline{0.1076} & \cellcolor{lightred}0.0662 \\ \hline
            mounted on & \textbf{0.0078} & \cellcolor{lightred}0 & \cellcolor{lightred}0 & \cellcolor{lightred}0 & \cellcolor{lightred}0 & \cellcolor{lightred}0 \\ \hline
            near & \textbf{0.0524} & \cellcolor{lightred}\underline{0.0481} & \cellcolor{lightred}0.0314 & \cellcolor{lightred}0.0240 & \cellcolor{lightred}0.0366 & \cellcolor{lightred}0.0195 \\ \hline
            of & 0.2689 & \cellcolor{lightred}0.2026 & \cellcolor{lightgreen}\underline{0.3438} & \cellcolor{lightgreen}0.3236 & \cellcolor{lightgreen}0.3092 & \cellcolor{lightgreen}\underline{0.3541} \\ \hline
            on & 0.0619 & \cellcolor{lightgreen}0.0701 & \cellcolor{lightgreen}\textbf{0.0970} & \cellcolor{lightgreen}\underline{0.0904} & \cellcolor{lightgreen}0.0884 & \cellcolor{lightgreen}0.0672 \\ \hline
            over & \textbf{0.0273} & \cellcolor{lightred}\underline{0.0186} & \cellcolor{lightred}0 & \cellcolor{lightred}0 & \cellcolor{lightred}0 & \cellcolor{lightred}0 \\ \hline
            parked on & 0.3508 & \cellcolor{lightgreen}\underline{0.3790} & \cellcolor{lightgreen}0.3711 & \cellcolor{lightred}0.3500 & \cellcolor{lightgreen}\textbf{0.3813} & \cellcolor{lightred}0.0169 \\ \hline
            riding & 0.3820 & \cellcolor{lightgreen}0.3836 & \cellcolor{lightgreen}\underline{0.4290} & \cellcolor{lightgreen}0.4203 & \cellcolor{lightgreen}0.4213 & \cellcolor{lightgreen}\textbf{0.4474} \\ \hline
            sitting on & 0.1647 & \cellcolor{lightgreen}\underline{0.1760} & \cellcolor{lightred}0.1427 & \cellcolor{lightgreen}\textbf{0.1813} & \cellcolor{lightred}0.1634 & \cellcolor{lightgreen}0.1742 \\ \hline
            standing on & \textbf{0.0683} & \cellcolor{lightred}\underline{0.0561} & \cellcolor{lightred}0.0155 & \cellcolor{lightred}0.0379 & \cellcolor{lightred}0.0384 & \cellcolor{lightred}0.0078 \\ \hline
            under & \underline{0.1281} & \cellcolor{lightred}0.1258 & \cellcolor{lightgreen}\textbf{0.1506} & \cellcolor{lightred}0.1241 & \cellcolor{lightred}0.1101 & \cellcolor{lightred}0.0928 \\ \hline
            using & \textbf{0.1263} & \cellcolor{lightred}\underline{0.0927} & \cellcolor{lightred}0 & \cellcolor{lightred}0 & \cellcolor{lightred}0 & \cellcolor{lightred}0 \\ \hline
            walking on & 0.1886 & \cellcolor{lightgreen}0.2113 & \cellcolor{lightgreen}\underline{0.2661} & \cellcolor{lightgreen}0.2272 & \cellcolor{lightgreen}0.2318 & \cellcolor{lightgreen}\textbf{0.2660} \\ \hline
            watching & 0.0934 & \cellcolor{lightgreen}\underline{0.1773} & \cellcolor{lightgreen}\textbf{0.2015} & \cellcolor{lightgreen}0.1571 & \cellcolor{lightgreen}0.1423 & \cellcolor{lightgreen}0.1682 \\ \hline
            wearing & 0.3886 & \cellcolor{lightgreen}\underline{0.4633} & \cellcolor{lightgreen}0.4546 & \cellcolor{lightgreen}0.4166 & \cellcolor{lightred}0.3568 & \cellcolor{lightgreen}\textbf{0.4871} \\ \hline
            wears & \underline{0.0450} & \cellcolor{lightred}0.0035 & \cellcolor{lightred}0.0270 & \cellcolor{lightred}0.0389 & \cellcolor{lightgreen}\textbf{0.1155} & \cellcolor{lightred}0.0017 \\ \hline
            with & 0.0230 & \cellcolor{lightgreen}0.0230 & \cellcolor{lightgreen}\underline{0.0388} & \cellcolor{lightgreen}0.0257 & \cellcolor{lightgreen}0.0331 & \cellcolor{lightgreen}\textbf{0.0415} \\ \hline
        \end{tabular}
    }
    \label{tab:AugPredRecall}
\end{table}


\subsection{Training from Synthetic Data}

\noindent Contrary to the findings with dataset augmentation, when training models from scratch, Stable Diffusion 2.0 images yielded slightly better performance than SGAdapter generated images. This is in line with expectations, as more realistic and higher quality images that more closely reflect real data provide better training information for models with no prior knowledge. Among the SGAdapter variants, Configuration 4 achieved the best results, occupying an halfway position - it produces images with a visual quality approaching Stable Diffusion 2.0, but with a less accurate structural representation compared to other SGAdapter configurations.

\noindent The Recall@K measures in Table~\ref{tab:Recall} confirm this pattern, showing that Stable Diffusion 2.0-generated images consistently produced higher results for both standard Recall and No Graph Constraint Recall metrics.

\begin{table*}[t]
    \centering
    \caption{\footnotesize Performance comparison of Recall and No Graph Constraint Recall metrics evaluated on the real data test set for CausalTDE models trained on the various synthetic datasets. Green cells indicate performance equal to or better than the baseline (synthetic data generated from Stable Diffusion 2.0), while red cells show performance degradation.}
    \scriptsize
    \begin{tabularx}{\linewidth}{|c|Y|Y|c|c|c|Y|Y|c|}
        \hline
        \multirow{2}{*}{\textbf{Model}} & \multicolumn{2}{c|}{\textbf{Attention Mask}} & \multicolumn{3}{c|}{\textbf{Recall}} &\multicolumn{3}{c|}{\makecell{\textbf{No Graph Constraint} \\ \textbf{Recall}}} \\
        \cline{2-3}\cline{4-9}
        & SGC-Att & S-Att & R@20 & R@50 & R@100 & NG-R@20 & NG-R@50 & NG-R@100 \\
        \hline
        \hline
        \makecell{Stable \\ Diffusion 2.0} & / & / & \underline{0.0074} & \textbf{0.0126} & \textbf{0.0186} & \textbf{0.0079} & \textbf{0.0135} & \textbf{0.0181}\\
        \hline
        \hline
        \makecell{SGAdapter\\Configuration 1} & \checkmark & \checkmark & \cellcolor{lightred}0.0063 & \cellcolor{lightred}0.0105 & \cellcolor{lightred}0.0153 & \cellcolor{lightred}0.0067 & \cellcolor{lightred}0.0106 & \cellcolor{lightred}0.0137 \\
        \hline
        \makecell{SGAdapter\\Configuration 2} & \checkmark & $\times$  & \cellcolor{lightred}0.0064 & \cellcolor{lightred}0.0103 & \cellcolor{lightred}0.0144 & \cellcolor{lightred}0.0072 & \cellcolor{lightred}0.0112 & \cellcolor{lightred}0.0142 \\
        \hline
        \makecell{SGAdapter\\Configuration 3} & $\times$ & \checkmark  & \cellcolor{lightred}0.0069 & \cellcolor{lightred}\underline{0.0112} & \cellcolor{lightred}0.0165 & \cellcolor{lightred}0.0072 & \cellcolor{lightred}0.0114 & \cellcolor{lightred}0.0148 \\
        \hline
        \makecell{SGAdapter\\Configuration 4} & $\times$ & $\times$ & \cellcolor{lightgreen}\textbf{0.0078} & \cellcolor{lightgreen}\textbf{0.0126} & \cellcolor{lightred}\underline{0.0185} & \cellcolor{lightred}\underline{0.0075} & \cellcolor{lightred}\underline{0.0124} & \cellcolor{lightred}\underline{0.0160}\\
        \hline
    \end{tabularx}
    \label{tab:Recall}
\end{table*}



\subsection{Qualitative results}

\begin{figure}
    \centering
    \includegraphics[width=1.\linewidth]{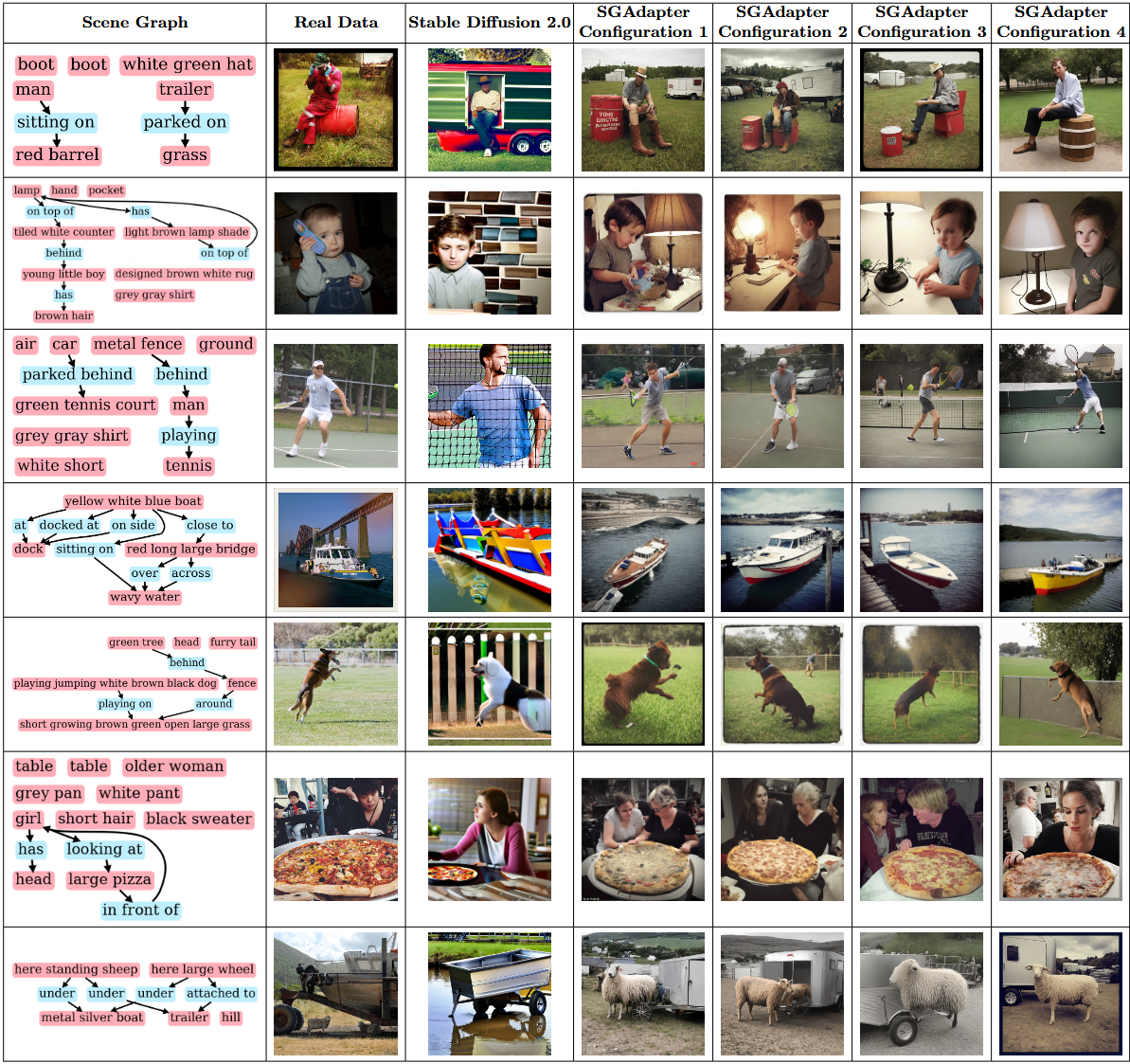}
    \caption{\footnotesize Visual comparison between the real images and the corresponding scene graphs, and the images generated with the various proposed models.}
    \label{fig:GeneratedImages}
\end{figure}

\noindent Table~\ref{fig:GeneratedImages} demonstrates that images generated with symbolic information exhibit enhanced structure, more accurate attribute assignment, and closer alignment with scene graph specifications. Stable Diffusion 2.0, while producing higher visual quality, frequently fails to differentiate multiple objects, incorrectly combines attributes, or inaccurately represents spatial relationships. The integration of SGAdapter clearly sacrifices some image quality, object realism, and definition to achieve improved semantic structure. SGAdapter Configuration 4 represents an intermediate solution, generating images with slightly reduced structural accuracy compared to other adapter configurations but somewhat improved visual quality. As scene graph complexity increases, Stable Diffusion-generated images show progressive degradation in quality and object differentiation, whereas SGAdapter-generated images maintain better object distinction despite occasional failures in representing complete scene graphs.

\noindent These observations align with the quantitative results: SGAdapter-generated images, despite appearing less coherent when evaluated in isolation, provide complementary semantic information during dataset augmentation. By enriching the real dataset with structured semantic content not effectively captured by pure diffusion models, these neuro-symbolically generated images enable superior performance improvements in downstream tasks despite their lower apparent visual quality.



\subsection{Limitations}
It is important to emphasize that SGAdapter was selected specifically for its efficiency as a lightweight adapter model that requires minimal training resources. However, unlike more computationally intensive alternatives that fundamentally alter the behavior of downstream image generators, SGAdapter makes relatively modest modifications. This deliberate design choice results in expected limitations: SGAdapter cannot accurately generate all predicates, which contributes to the observed reduction in image quality.

\noindent The experimental framework also introduces inherent constraints through the selected Scene Graph Generation task and the CausalTDE model with MotifsNet backbone. SGG does not fully leverage the potential improvements offered by SG-to-Image generation, as it neglects attributes and fails to differentiate between multiple instances of the same object type. The SGG model selection prioritized training efficiency over complexity, resulting in a model that relies less on precise image structure and more on statistical predicate-object co-occurrence patterns compared to more sophisticated but computationally demanding alternatives.

\section{Conclusions}
\noindent This work investigated whether integrating Neuro-Symbolic approaches could address limitations in synthetic dataset generation, focusing on Scene Graph-conditioned image generation for SGG tasks. Experiments with SGAdapter and Stable Diffusion 2.0 revealed that for dataset augmentation, NeSy-generated images produced significantly better results than conventional generation approaches, with up to 2.59\% improvement in Recall metrics. Although NeSy-generated images exhibited lower perceptual quality when evaluated independently, they contributed complementary structural information when combined with real data, enriching the dataset with semantic representations not effectively encoded by conventional diffusion models. The explicit relational constraints imposed by scene graphs enhance model learning despite the decreased visual realism.

\noindent Future research should address current limitations by implementing more sophisticated NeSy architectures, integrating these generators into advanced dataset creation pipelines, and expanding their application to diverse reasoning tasks beyond Scene Graph Generation.

\section*{Acknowledgements}
\noindent This paper is supported by the FAIR (Future Artificial Intelligence Research) project, funded by the NextGenerationEU program within the PNRR-PE-AI scheme (M4C2, investment 1.3, line on Artificial Intelligence).

\bibliography{references}
\bibliographystyle{elsarticle-num}

\end{document}